\documentclass[letterpaper, 10 pt, conference]{ieeeconf}  

\IEEEoverridecommandlockouts                              

\usepackage{graphics} 
\usepackage{epsfig} 
\usepackage{mathptmx} 
\usepackage{times} 
\usepackage{amsmath} 
\usepackage{amssymb}  
\usepackage{hyperref}
\usepackage[capitalize]{cleveref}
\usepackage[per-mode=symbol]{siunitx}
\usepackage{todonotes}
\usepackage{pgfplots}
\usepackage[protrusion=true,expansion=true]{microtype} 
\usepackage[keeplastbox]{flushend}
\usepgfplotslibrary{groupplots}
\usetikzlibrary{positioning}
\newcommand{\argmax}{\operatornamewithlimits{arg\,max}}

\usepackage[backend=bibtex,style=ieee, maxcitenames=2, mincitenames=1]{biblatex}
\AtEveryBibitem{
 	\clearfield{url}  
 	\clearfield{doi}  
\ifentrytype{inproceedings}{
 	\clearlist{address}
 	\clearlist{publisher}
 	\clearname{editor}
 	\clearlist{organization}
 	\clearfield{pages}  
 	\clearlist{location}
 	\clearfield{volume}
 }{}
 }

\AtBeginBibliography{\small}

\title{\LARGE \bf
Safe Reinforcement Learning with Scene Decomposition \\ for Navigating Complex Urban Environments
}

\author{Maxime Bouton,$^1$ Alireza Nakhaei,$^2$ Kikuo Fujimura,$^2$ and Mykel J. Kochenderfer$^1$%
    \thanks{*This work was supported by the Honda Research Institute.}
    \thanks{$^{1}$ Maxime Bouton and Mykel J. Kochenderfer are with the Department of Aeronautics and Astronautics, Stanford University, Stanford CA 94305, USA,
            {\tt \{boutonm,mykel\}@stanford.edu}.}%
    \thanks{$^{2}$ Alireza Nakhaei and Kikuo Fujimura are with the Honda Research Institute, 375 Ravendale Dr., Mountain View, CA 94043, USA, 
            {\tt {anakhaei,kfujimura}@hra.com}.}%
}

\begin{document}

\maketitle
\thispagestyle{empty}
\pagestyle{empty}

\begin{abstract}
Navigating urban environments represents a complex task for automated vehicles. They must reach their goal safely and efficiently while considering a multitude of traffic participants.
We propose a modular decision making algorithm to autonomously navigate intersections, addressing challenges of existing rule-based and reinforcement learning (RL) approaches.
We first present a safe RL algorithm relying on a model-checker to ensure safety guarantees. To make the decision strategy robust to perception errors and occlusions, we introduce a belief update technique using a learning based approach.
Finally, we use a scene decomposition approach to scale our algorithm to environments with multiple traffic participants. We empirically demonstrate that our algorithm outperforms rule-based methods and reinforcement learning techniques on a complex intersection scenario.
\end{abstract}

\section{INTRODUCTION}

\begin{figure}[t!]
    \centering
    \includegraphics[width=0.8\columnwidth]{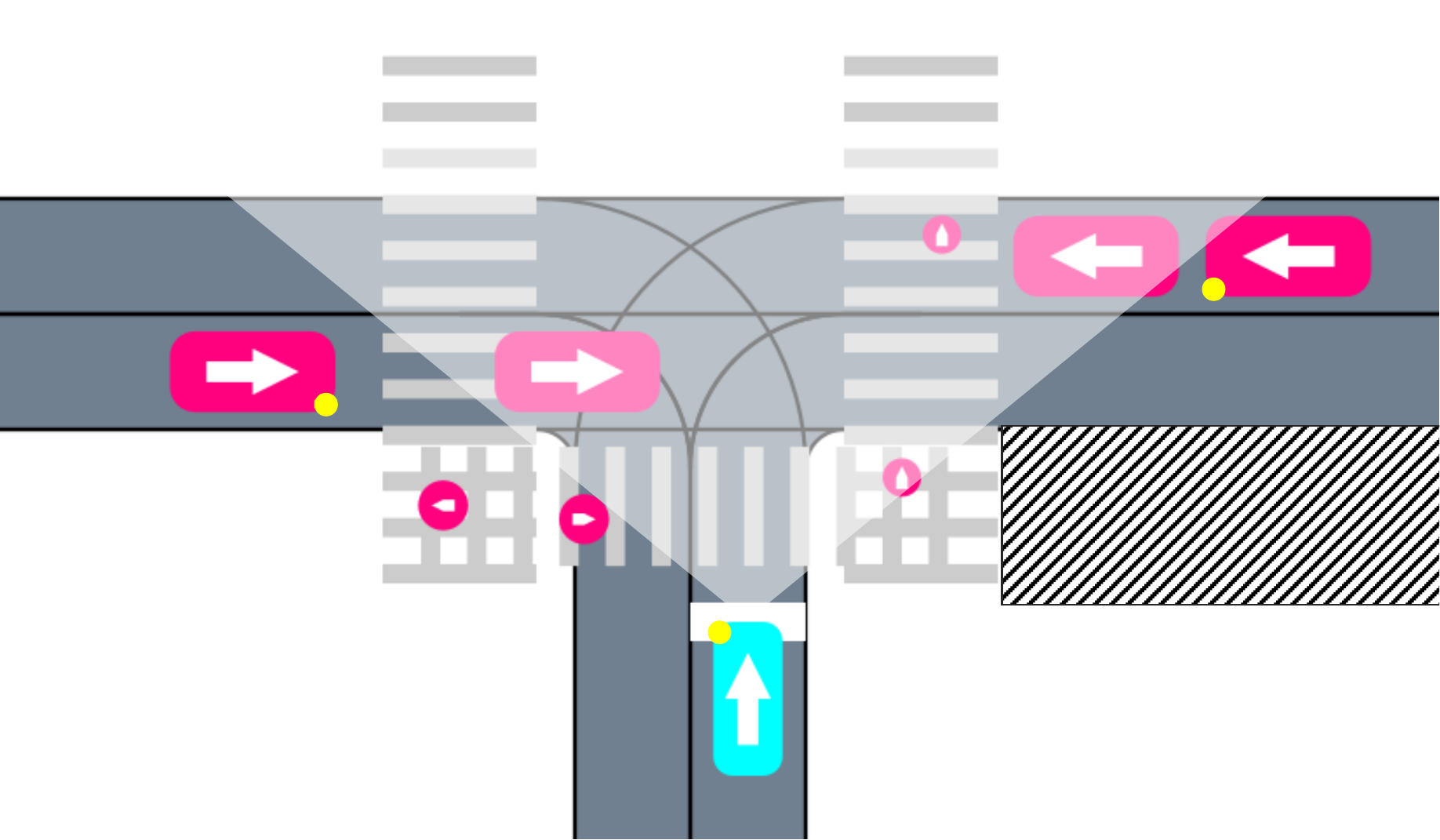}
    \caption{Illustration of a typical driving scene handled by our algorithm. Pedestrians and cars might be occluded by obstacles (hatched), and exhibit complex interactive behavior. The ego vehicle (blue) must control its longitudinal acceleration along a given path to achieve a left turn safely and efficiently.}
    \label{fig:scenario}
\end{figure}

Automated driving has the potential to significantly improve safety. Although major progress in enabling this technology has been made in recent years, autonomously navigating urban environments efficiently and reliably remains challenging. At urban intersections, vehicles must navigate among both cars and pedestrians, using on board perception systems that give noisy estimates of the location and velocity of others  on the road and are sensitive to occlusions (\cref{fig:scenario}). Autonomously navigating urban intersections requires algorithms that reason about interactions between traffic participants with limited information. 



Engineering a rule-based strategy to navigate such an environment would require anticipating the vast space of possible situations. A common heuristic strategy is to use a threshold on the time to collision~\cite{alonso2011}. Such an approach performs well in simple scenarios but does not take into account sensor uncertainty and is unlikely to scale to complex environments. Alternatively, previous work suggests modeling the problem as a partially observable Markov decision process (POMDP)~\cite{brechtel2014, bouton2017, hubmann2018}. POMDPs provide a principled framework to model uncertainty of other drivers' intent through latent variables, as well as integrating perception and planning~\cite{bouton2017, hubmann2018}. However, these methods are often difficult to scale in environments with multiple road users. 

Reinforcement learning (RL) has been proposed as a way to automatically generate effective behaviors.  RL has been applied to autonomous braking strategies at crosswalks~\cite{chae2017}, lane changing policies~\cite{wang2018}, and intersection navigation~\cite{tram2018, isele2018}. \citeauthor{tram2018} propose a deep reinforcement learning approach with recurrent neural networks to learn how to navigate intersections with multiple vehicles with changing behaviors~\cite{tram2018}. \citeauthor{isele2018} address a scenario with sensor occlusions by representing the state as an occupancy grid and learning a policy through Q-learning~\cite{isele2018}. Although both approaches show promising results and efficient policies, they often fail at providing safety guarantees.

To enforce safety in decision making algorithms, different techniques have been proposed~\cite{garcia2015}. Conservative rule-based strategies based on traffic rules and short term predictions can be used to constrain the action of a reinforcement learning agent~\cite{mukadam2017, mirchevska2018}. Formal methods can be used to derive shielding mechanisms in a more systematic way~\cite{alshiekh2018}. These techniques rely on a model of the environment and provide strong safety guarantees. They often require an abstraction of the environment that might be difficult to design.

In this work, we propose a decision making framework to navigate urban intersections. Our algorithm integrates concepts from the POMDP planning, reinforcement learning, and model checking literature to address some of the challenges of autonomously navigating urban environments pertaining to safety, efficiency, robustness, and scalability. We first present the combination of a model-checker and a reinforcement learning policy to derive efficient policies with probabilistic safety guarantees. Then, we introduce a new belief update approach that uses an ensemble of neural networks. The output of these networks are used as input for the RL algorithm.
Finally, a scene decomposition method enables us to scale our approach to a large number of agents and take advantage of local interactions between traffic participants. We provide an ablation study of the different components of our decision making system to illustrate the benefit of each on five evaluation scenarios. Simulation results show that our algorithm outperforms RL policies and an engineered rule-based method.
\section{BACKGROUND}

\subsection{Reinforcement Learning}

Sequential decision making problems can be modeled by a Markov decision process (MDP). MDPs are mathematical frameworks defined by the tuple $(\mathcal{S}, \mathcal{A}, T, R, \gamma)$, where $\mathcal{S}$ is a state space, $\mathcal{A}$ is an action space, $T$ is a transition model, $R$ is a reward function, and $\gamma$ is a discount factor. The agent takes an action $a$ at a given state $s$, and the environment evolves to a state $s'$ with a probability $T(s, a, s') = \Pr(s \mid s',a)$. After every transition, the agent receives a reward $r = R(s, a)$ for taking action $a$ in state $s$. The action $a$ is chosen according to a policy $\pi: \mathcal{S} \rightarrow \mathcal{A}$.
We call the state-action utility of a policy $\pi$ the quantity $Q^\pi(s, a) = E[\sum_{t=0}^\infty \gamma^t r_t | s_0 = s]$. This quantity represents the discounted accumulated reward obtained by the agent when taking action $a$ from state $s$ and then following policy $\pi$. The optimal state-action utility verifies the Bellman equation~\cite{bellman1957}:
\begin{equation}
    Q^*(s, a) = R(s, a) + \gamma \sum_{s'}T(s, a, s')\max_a Q^*(s', a)
    \label{eq:bellman}
\end{equation}

In MDPs with finite state space and action space, \cref{eq:bellman} can be solved using value iteration. However, when the state space is continuous and high dimensional, approximation methods must be used~\cite{dmu}. In recent years, deep reinforcement learning algorithms have been shown to find efficient policies in very large MDPs. In deep reinforcement learning, the state-action value function is represented by a neural network: $Q(s, a; \theta)$ where $\theta$ encodes the weights of the network. 
The solution to \cref{eq:bellman}, can be approximated by the network minimizing the following loss function:
\begin{equation}
	J(\theta) = \mathbb{E}_{s'}[(r + \gamma\max_{a'}Q(s', a'; \theta) - Q(s, a; \theta))^2]
\end{equation}
Given an experience sample $(s, a, r, s')$, the weights are updated as follows:
\begin{equation}
	\theta \leftarrow \theta + \alpha (r+ \gamma\max_{a'}Q(s', a'; \theta) - Q(s, a; \theta))\nabla_\theta Q(s,a;\theta)
\end{equation}
where $\alpha$ is the learning rate, a hyperparameter of the algorithm. \citeauthor{mnih2015} proposed several innovations to improve network training, such as the use of a target network and experience replay, which lead to a scalable reinforcement learning solver known as deep Q-learning (DQN)~\cite{mnih2015}.

\subsection{State Uncertainty}

MDP models assume that the agent can observe the true state of the environment perfectly. However, in autonomous driving scenarios, the ego vehicle receives imperfect observations of the environment. Hence, the autonomous driving problem is inherently a partially observable Markov decision process (POMDP). In a POMDP, the agent represents its knowledge of the environment with a belief state $b: \mathcal{S} \rightarrow [0,1]$ such that $b(s)$ is the probability of being in state $s$. At every time step, the agent receives an observation and updates its belief. Algorithms that update a current belief given observations are referred to as belief updaters, or filters in the tracking literature. 

Solving decision making problems in the belief space is generally intractable. Instead, one can use the QMDP approximation~\cite{littman1995}:
\begin{equation}
    Q(b, a) = \sum_s Q_{\text{MDP}}(s, a)b(s)
\end{equation}
where $Q_{\text{MDP}}$ is the solution to the problem considered as an MDP. Such approximation assumes that the state will be perfectly observable at the next time step. The QMDP method over approximates the true belief-state value function. As a consequence, an agent following such a policy will not take information gathering actions. We argue that for the problem of interest, information gathering does not need to be incentivized and will occur naturally as the agent moves towards the goal. Experiments show that the QMDP approximation suffices to make our algorithm robust to state uncertainty.

\section{PROPOSED APPROACH}

\begin{figure}
    \centering
    \includegraphics[width=\columnwidth]{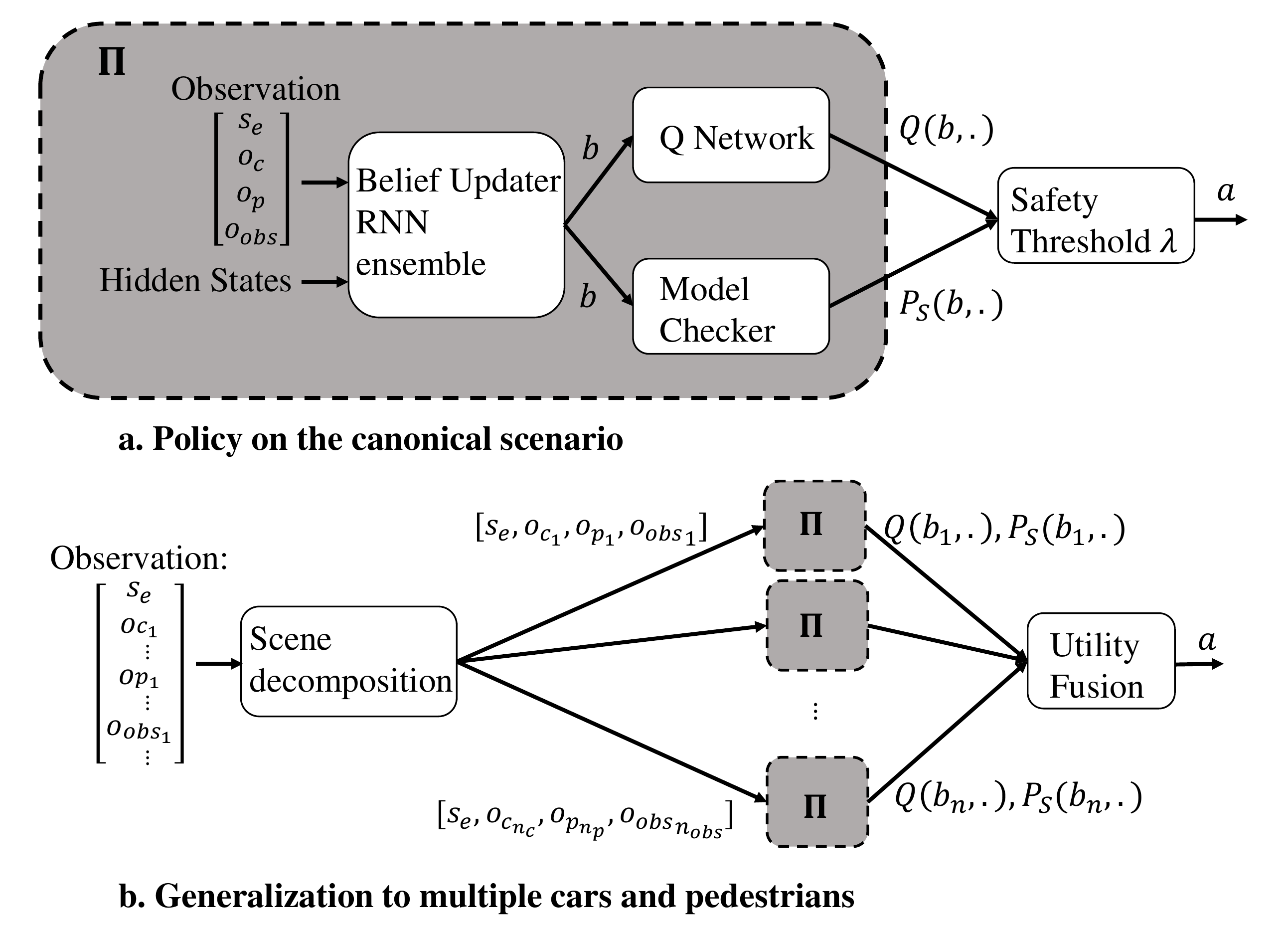}
    \caption{Flow diagrams of our decision making algorithm. The top figure shows the policy acting on the canonical scenario, integrating the belief updater, model checker, and RL agent. The bottom figure illustrates the scene decomposition method used to generalize the previous policy.}
    \label{fig:flowchart}
\end{figure}

This section describes the different components of our approach, illustrated in \cref{fig:flowchart}. We first explain how to model the intersection navigation problem. We then present our safe reinforcement learning algorithm and learned belief updater on a canonical scenario involving only the ego vehicle, a single other car, and a pedestrian. Finally, we demonstrate how to use the solution to this canonical scenario to navigate intersections with a multitude of cars and pedestrians. The hyperparameters for each of the methods presented are available in our code base\footnote{\url{https://github.com/sisl/AutomotiveSafeRL}}.

\subsection{Modeling Intersection Scenarios}\label{sec:modeling}

The urban intersections scenario is modeled as a Markov decision process. A simulation environment\footnote{The implementation is available at \url{https://github.com/sisl/AutomotivePOMDPs.jl}} is derived from the MDP formulation by sampling the transition model. In addition, we simulate perception error during evaluation, making our model a POMDP. 

\paragraph{State} 
We define the state of a traffic participant $c$ to be $s_c=(x, y,\theta, v)$ which is its position, heading, and its longitudinal velocity. We used a Cartesian frame with the origin at the center of the intersection to define the position. Fixed obstacles are defined by their positions and their sizes: $s_{obs} = (x, y,\theta, l, w)$ where $l$ is the length and $w$ the width of the obstacle. The global state ${s = (s_{\text{ego}}, s_{\text{c}_{1:n_c}}, s_{\text{p}_{1:n_p}}, s_{\text{o}_{1:n_o}})}$ can be described as follows:
\begin{itemize}
	\item $s_{\text{ego}}$ represents the physical state of the ego vehicle 
	\item $s_{\text{c}_i}$ represents the physical state of the $i$-th car in the environment with $i=1\ldots n_{c}$ where $n_{c}$ is the number of cars present in the environment
	\item $s_{\text{p}_i}$ represents the physical state of the $i$-th pedestrian in the environment with $i=1\ldots n_{p}$ where $n_{p}$ is the number of pedestrians present in the environment.
	\item $s_{\text{o}_i}$ is the pose of the $i$-th obstacle present in the environment with $i=1\ldots n_{o}$ where $n_{o}$ is the number of fixed obstacles present in the environment.
\end{itemize}

In addition, we add an extra state variable, $s_{\text{absent}}$, to model a potential incoming traffic participant that is not present in the scene. Contrary to previous work, this description of the state takes into account any obstacle configuration and any type of entity (pedestrian or car) in the problem formulation. Other POMDP approaches consider adding a latent variable to each of the traffic participants to describe their behavior \cite{hubmann2018, bouton2017}. In this paper, uncertainty about other traffic participants' behavior is captured by the transition model and state uncertainty only takes into account sensor limitations. However, the proposed framework could be extended to model the intentions of drivers and pedestrians.

\paragraph{Action Space} The ego vehicle controls its acceleration along a given path by choosing an acceleration level among the set: $\{\SI{-4}{\meter\per\second\squared}, \SI{-2}{\meter\per\second\squared}, \SI{0}{\meter\per\second\squared}, \SI{2}{\meter\per\second\squared}\}$ which corresponds to a comfortable driving.

\paragraph{Transition} The transition model is designed to capture interaction between traffic participants. For a state with a single car $c$ and a single pedestrian $p$, we can factorize the transition model as follows:
\begin{equation}
\Pr(s' \mid s, a) = P_{\text{ego}}(s'_{\text{ego}} \mid s_{\text{ego}}, a)P_c(s'_c \mid s )P_p(s'_p \mid s)
\end{equation}
$P_{\text{ego}}$ represents the dynamics of the ego vehicle and is modeled by a deterministic point mass dynamic. $P_c$ and $P_p$ represent the model of the other car and the pedestrian respectively. The car follows a rule-based policy described in \cref{sec:experiment} and the pedestrian follows a time to collision policy to decide whether it is safe to cross the street. Hence, the actions of the car and the pedestrian depend on their respective state as well as the state of the ego vehicle. To describe uncertainty in the behavior of other vehicles, a Gaussian noise with standard deviation \SI{2}{\meter\per\second\squared} is added to the output of the rule-based policy. $P_c$ represents the model of the pedestrian. In addition, new cars and pedestrians can appear in the scene with a constant probability of appearance at each time step. They appear at the beginning of any lane or crosswalk randomly with a random velocity.

\paragraph{Observation} The ego vehicle receives a noisy observation of the state according to the following sensor model:
\begin{itemize}
    \item The position measurement follows a Gaussian distribution centered around the ground truth with standard deviation $\sigma_p$ growing linearly with the distance to the target.
    \item The velocity measurement follows the same model with $\sigma_v$ growing linearly with the distance to the target.
    \item There is a false negative rate of \num{0.1} and a false positive rate of \num{0.1} if no targets are visible.
    \item If a target is occluded, it cannot be detected. In the simulated environment we compute occlusion by a ray tracing technique: if the segment connecting the front of the ego vehicle and the target intersects with the obstacle, then the target is occluded.
\end{itemize}
The observation model is used for simulation purposes. The safe-RL algorithm that we will present in the next section derives a policy in a fully observable environment.

Such mathematical formulation provides a principle framework to model multiple vehicles interacting, uncertain behaviors, and limited perception of the environment. However, solving for the optimal strategy to navigate in such an environment would be intractable due to an exponential increase in the number of states with the number of agents considered. For this reason, we will first focus on a sub-task in a simpler scenario, using offline methods. We consider a canonical scenario involving the ego vehicle, a single other car, and a single other pedestrian. This scenario is sufficient to capture complex interactions such as the other car yielding to a crossing pedestrian. The generalization to environments with multiple agents and obstacles is performed online and is addressed in \cref{sec:decomposition}.

\subsection{Safe Reinforcement Learning}

\begin{figure*}[t!]
    \centering
    \includegraphics[width=2\columnwidth]{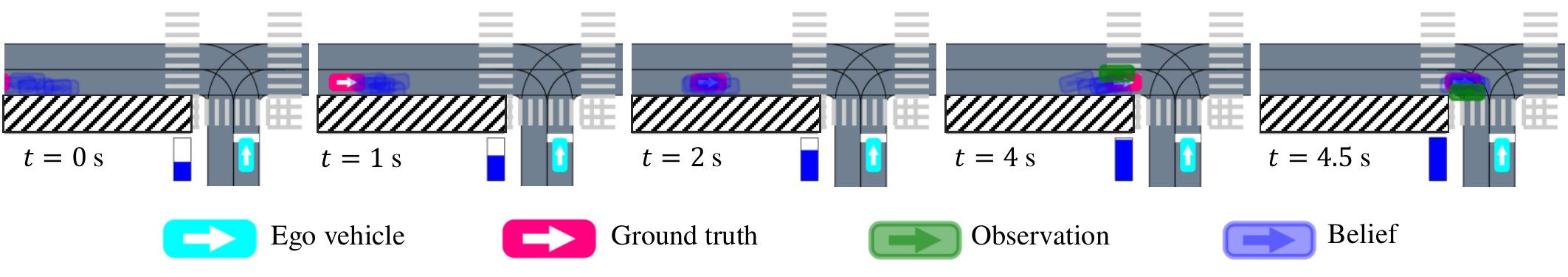}
    \caption{Example of a trajectory where the ensemble of RNNs tracks a car through an occluded area. The blue gauge represents the predicted probability of presence of a vehicle behind an obstacle. An empty gauge corresponds to a probability of 0 and a full gauge to a probability of 1. In this example, the ego vehicle is still at the stop line in order to demonstrate the ability of the belief updater to track through obstacles during a long time period.}
    \label{fig:belief}
\end{figure*}

Deep reinforcement learning methods could derive efficient policies for the proposed scenario but do not provide any safety guarantees. For safety critical applications such as autonomous driving, a way to benefit from the flexibility and scalability of deep RL is to constrain the action space to safe actions~\cite{bouton2018reinforcement, garcia2015}. 
In this work, we use a probabilistic model checker to compute the probability of reaching the goal safely for each state-action pair prior to learning a policy. Safe actions are then identified by applying a threshold on the probability of success at a given state-action pair, similar to the method used by \citeauthor{bouton2018reinforcement}~\cite{bouton2018reinforcement}. The threshold is a user defined parameter which represents the minimum desired probability of success. 

The model checker relies on the value iteration algorithm and requires a discretization of the state space and the full specification of the transition model~\cite{baier2008}. To keep the representation tractable, the state space is limited to longitudinal positions and velocities for the three agents, a variable indicating the car's lane and a variable indicating the pedestrian's crosswalk. The pedestrian can be on any of the three crosswalks considered and travels in both direction (making it six values for the pedestrian lane). The car can drive in any of the lanes present on \cref{fig:scenario}.  By choosing resolutions of \SI{2}{\meter} for the position and \SI{2}{\meter\per\second} for the velocity of each agent, the number of states is approximately \num{23e6}. Given this discrete representation, the probability of reaching the goal safely, $P_S(s, a)$, can be computed offline using parallel value iteration.

A safety threshold $\lambda$ is used to constrain the agent to take actions inside the set $\mathcal{A}_{\text{safe}}=\{ a \mid P_S(s, a) > \lambda \}$. In cases where the set is empty, the agent then executes the safest possible action given by the model checker. In cases where $\mathcal{A}(s)$ is not empty, the agent can choose any actions within the set. We train an RL agent to choose the best actions among the possible safe actions as follows:
\begin{equation}
    \pi_{\text{safe}}(s) = \begin{cases} 
             \argmax_{ a\in \mathcal{A}_{\text{safe}}(s)} Q(s, a)& \text{if} \mathcal{A}_{\text{safe}}(s)\neq\emptyset\\
             \argmax_a P_S(s, a) & \text{if} \mathcal{A}_{\text{safe}}(s)=\emptyset \\
            
    \end{cases}
\end{equation}
This constrained action selection strategy transfers the safety guarantees of the model checker to the RL agent~\cite{bouton2018reinforcement}.

The probabilistic model checker enables identifying safe actions and bounding the actions of the RL agent. However, the ego vehicle must also reach the goal as fast as possible. To optimize for this objective, we define a simple reward function that assigns a value of \num{1} to goal states. Collision states are already avoided thanks to the constrained exploration. The policy is trained using deep Q-learning with a constrained action space to enforce safety.  The training environment is a continuous state space, simulated environment, following the model described in \cref{sec:modeling}, with only one other car and one pedestrian and perfect observation. The policy is modeled by a feedforward neural network with four layers of \num{32} nodes and ReLU activations. The input to the network is a twelve dimensional vector with the positions (2D), longitudinal velocity, and heading of the ego vehicle, the car, and the pedestrian. Future work could investigate more complex reward design with terms for passenger comfort or social behavior. 

The safe RL algorithm provides a policy that is trained on the canonical scenario optimizing for time efficiency under safety constraint imposed by the model checker. Computing the probability of collision and training the policy assumes an MDP model of the environment where the position and velocity of the road users are fully observable.

\subsection{Ensemble Belief Updater}



In this section we present a belief updater capturing state uncertainty. A belief updater is generally an algorithm that updates the new belief given the old belief, and the current observation. Although perception systems already use filtering algorithms, their outputs are still noisy and may result in false positive and false negative. In this work, the belief updater has sole purpose to integrate perception error in the planning algorithm. It is assumed that a perception system is processing raw sensor data beforehand and provides a structured representation of the environment. Our belief updater assumes perfect data association. Each entity is associated to an ID that is consistent through the whole trajectory. 

Classic belief updaters, such as particle filters, could be used to track occluded object and smooth perception errors~\cite{galceran2015}. However, most of these algorithms rely on some model of the environment. In this paper, we experiment with a new learning-based approach to represent belief states. The main motivation is to capture environment specific parameters, such as the probability of an entity appearing in an occluded area, as well as to learn scenario specific dynamics, such as a car yielding to a pedestrian, which cannot be easily modeled.

Our belief update algorithm consists of an ensemble of recurrent neural networks (RNN). The hidden state of each network is responsible for keeping track of the observation history. The input of each network is an observation vector, and the output is the predicted ground truth position of the car and the pedestrian as well as a probability of presence. The input to the RNN is a sixteen dimension vector encoding the observed state of the ego vehicle, pedestrian, other car, and obstacle. The prediction is a ten dimension vector: the predicted state of the car and pedestrian, and their probability of presence. 
The recurrent neural network is trained using gradient descent on the mean squared error between ground truth trajectories and predicted trajectories. 

In order to make the prediction more robust, we used an ensemble method~\cite{sollich1995}. Instead of training a single network, $k$ different networks are trained on a different portion of the dataset. Those randomly initialized networks will converge to different local optima. As a result, they will each give different predictions for a given input. We use an ensemble of five networks and the five predictions represents the belief of the agent. The use of an ensemble of RNNs emulates the role of a particle filter similarly to the approach presented by \citeauthor{igl2018}~\cite{igl2018}. 

To train the network, we generated a synthetic dataset with the same simulation environment used for training the RL policy. This scenario involves one ego vehicle, one other car, and one pedestrian, as well as one obstacle randomly placed on the side of the road. We generated \num{3000} trajectories of \num{400} time steps of \SI{0.1}{\second}. The belief updater is trained to handle up to a certain level of perception noise described in \cref{sec:experiment}.

In \cref{fig:belief} we can see an example of our belief updater predicting the state of a car behind an occluded area. The blue trace represents the predictions of the five RNNs. We can see that they all give different predictions, enhancing the robustness of the estimation. As the time increases, the probability of presence of a car behind the obstacle increases since our model assumes a constant rate of appearance at every time step. The network is able to infer that cars can appear at the beginning of the road because it has experienced such cases in the training set. Finally, on the last two frames the car is observed (green trace) and the predictions converge to a more accurate estimation. 

It is important to realize that standard particle filters could have been used since our environment is simulated and the model is known. However, using a learning based-method could help capture unknown parameters such as the probability of the appearance of a car, or pedestrian, as well as scenario specific dynamics, such as a car braking strongly to yield to a pedestrian. Experiments in \cref{sec:experiment} show that our ensemble of RNNs provides robustness to perception errors. Further work is needed to benchmark such approach against standard tracking algorithms and evaluate the number of RNNs to use in the ensemble. 


\subsection{Online Scene Decomposition}\label{sec:decomposition}

Thus far, we presented our approach on a scenario involving only three traffic participants. In this section we will demonstrate how to generalize the safe RL policy, and the belief updater, to situations with multiple cars and pedestrians through the use of scene decomposition. Decomposition methods have been used in the past to approximate the solution of large decision making problems~\cite{chryssanthacopoulos2012decomposition, wray2017}. They consist of combining the utility functions associated with simple tasks to approximate the solution to a more complex task. Previous work proposed an agent-wise decomposition where each target to avoid is considered independently~\cite{chryssanthacopoulos2012decomposition}.

In our intersection scenario, the simple task consists of navigating the intersection assuming only one car, one pedestrian, and one obstacle can be present. The motivation for using a canonical scenario with three agents is to capture interactions between cars and pedestrians. In the presence of multiple cars and pedestrians, the global belief, $b = (b_{\text{ego}}, b_{\text{c}_1}, \ldots, b_{\text{p}_1}, \ldots, b_{\text{o}_1})$, can be decomposed into multiple instances of the canonical scenario. That is:
\begin{equation}
    b = \{(b_{\text{ego}}, b_{\text{c}_1}, b_{\text{p}_1}, b_{\text{o}_1}),  \ldots, (b_{\text{ego}}, b_{\text{c}_1}, b_{\text{p}_1}, b_{\text{o}_1})\}
\end{equation}
At each time step, we will have $n_{c} \times n_{p} \times n_{obs}$ canonical beliefs. Since $n_c$ and $n_p$ are not known in practice due to sensor occlusions, the global state variable is augmented with an additional car and pedestrian observed as absent. This allows the agent to always assume that there is at least one agent that might appear from an occluded area.  

Once the belief state is decomposed into canonical belief states, we perform the following approximation:
\begin{align}
    P_S(b, a) &= \min_i P_S(b_i, a) \\
    Q(b, a) &= \min_i Q(b_i, a)
\end{align}
This operation takes into account the canonical belief with the worst probability of success and the worst utility. Entities that are far from the ego car and present very little risk will be associated to a higher utility and higher probability of success and will be ruled out from the decision. 

The computational cost of the scene decomposition method grows linearly with the number of cars or pedestrians considered. Once the belief state decomposed, a policy call requires evaluating $Q$ and $P_S$ $n_{c} \times n_{p} \times n_{obs}$ times online. Since $Q$ and $P_S$ are computed offline, evaluating them at a given belief point involves a pass forward through a neural network and a table query which is relatively fast, as illustrated in \cref{fig:comptime}. In our current implementation, we used a decision step of \SI{0.1}{\second} so the policy call would become slower than real time after \num{7} cars and one pedestrian. In practice, the decision step could be reasonably increased up to \SI{0.5}{\second}, and the different calls to $Q$ and $P_S$ could be parallelized. Updating the belief using the ensemble of networks stayed below \SI{100}{\milli\second}.

The use of $\min$ to combine individual utilities could be overly conservative in crowded environments. Increasing the size of the canonical scenario could help taking into account more complex interaction. This fact highlights a computational trade-off between the number of traffic participants considered jointly or separately. Previous work has studied algorithms to bridge the gap between approximated solutions obtained from decomposition methods and the solution to the problem considering every traffic participants~\cite{bouton2018utility}.


\begin{figure}
    \centering
\begin{tikzpicture}[]
\begin{axis}[
legend pos={north west},
width=8cm,
height=5cm,
xmin = {1},
grid=both,
xlabel = {Number of cars},
ylabel = {Time (\SI{}{\milli\second})}
]
\addplot+[style={thick, black}, mark options={fill=black}]
coordinates {
(1.0, 16.9812)
(2.0, 32.0119)
(3.0, 49.743)
(4.0, 59.8401)
(5.0, 73.7714)
(6.0, 96.19)
(7.0, 103.81)
(8.0, 114.49)
(9.0, 118.77)
(10.0, 139.9)
};
\addlegendentry{Safe RL Policy}
\addplot+[style={thick, gray}, mark options={fill=gray}]
coordinates {
(1.0, 24.4736)
(2.0, 27.8628)
(3.0, 28.9514)
(4.0, 35.4398)
(5.0, 46.0694)
(6.0, 55.98)
(7.0, 65.9428)
(8.0, 65.44)
(9.0, 71.3)
(10.0, 76.128)
};
\addlegendentry{Ensemble Belief Update}
\end{axis}
\end{tikzpicture}
    \caption{Online computation time of calling the safe RL policy and the belief updater. We fixed the number of pedestrians to one and varied the number of cars present in the environment.}
    \label{fig:comptime}
\end{figure}

        

    

\section{EXPERIMENTS}\label{sec:experiment}

\begin{figure*}[t]
    \centering
    \includegraphics[width=2\columnwidth]{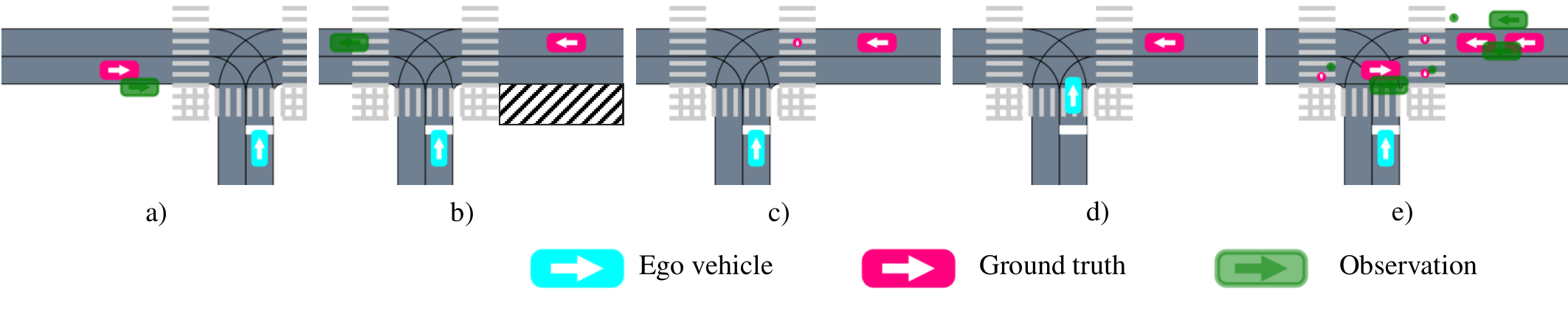}
    \caption{Examples of scenes from our five evaluation scenarios. In scenario a, b, and e, the ego vehicle has limited sensing capabilities. The green trace in scenario b shows an example of false positive detection.}
    \label{fig:eval-scenario}
\end{figure*}

We evaluated our algorithm on a simulated environment following the model described in \cref{sec:modeling}.
To measure the performance of our algorithm, we designed five different scenarios and measured the number of time steps to reach the goal as well as the number of collisions. We ran \num{10000} simulations and measured the average number of steps taken to reach the goal as well as the average number of collisions observed in simulation. The results are presented in \cref{fig:barplots}. 

\subsection{Evaluation Scenarios}

The first two scenarios, a and b, evaluate the benefit of using a belief updater. Scenarios c and d evaluate the benefit of having a three agent formulation as the canonical scenario. Finally, scenario e evaluates the scalability of our algorithm. \cref{fig:eval-scenario} shows a situation from each of those scenarios. Across the scenarios, we varied the sensor performances, the presence of obstacles, and the number of cars and pedestrians present in the scene. For each of them, the ego vehicle starts with no velocity, waiting at a T shape intersection with no stop signs. 

In all scenarios but c and d, the perception error is modeled by a Gaussian distribution around the position and velocity measurement with false positive and false negative rates. The standard deviations are \SI{0.5}{\meter} and \SI{0.5}{\meter \per\second} for the position and velocity measurements. We used a false positive rate of \num{0.1} and a false negative rate of \num{0.1}. The perception system is also sensitive to physical obstacles such as buildings. We used ray tracing to check if a traffic participant is occluded or not. If it is occluded, then no observation is received by the ego vehicle.

\paragraph{Perception Noise} We consider only one other car going straight and coming from the right or the left with equal probability. The car is initialized with a random longitudinal position along its route and a random longitudinal velocity between \SI{0}{\meter\per\second} and \SI{8}{\meter\per\second}. We can see from \cref{fig:barplots} that the five policies take similar times to reach the goal.

\paragraph{Occlusions} This scenario is the same as described above except that a physical obstacle is present on the side of the road, which blocks the sensing capabilities of the vehicle. The initial scene is generated such that the incoming car is always occluded by the obstacle.

The next two scenarios have been designed in order to evaluate the ability of our decision making system to capture interactions between traffic participants. In both scenarios we assume that the ego vehicle has perfect sensing capabilities.

\paragraph{Car and pedestrian interaction} In this scenario, we consider one other car and one other pedestrian interacting with each other. The strategy of the other car is to yield to the pedestrian if it is crossing or preparing to cross. The pedestrian is initialized on one of the three crosswalks randomly, with a random longitudinal position and a random velocity between \SI{0}{\meter\per\second} and \SI{2}{\meter\per\second}. The other car is initialized with a random route, a random longitudinal position, and a random velocity.

\paragraph{Ego and other car interaction} Only one other car, and no pedestrian is considered. The other vehicle is coming from the right and must perform a left turn. If the ego vehicle starts engaging in the intersection, then the other vehicle should yield to the ego vehicle. This scenario has been designed to evaluate the ability of our policy to exploit this interaction. 

\paragraph{Multiple cars and pedestrians} In this scenario, we evaluate the scalability of our approach. A flow of cars and pedestrians is generated with a probability of appearance of \num{0.1} at every time step. We also added observation noise and physical obstacles randomly initialized on the left or right of the ego vehicle. An example scene from scenario e is illustrated in \cref{fig:scenario}.

\subsection{Baseline Policies}

We compared our algorithm's performance against a rule-based method as well as an RL policy. 

\textbf{Rule-based policy:} This policy is a hand-engineered rule-based strategy. It is the same policy followed by human cars in our simulation environment. The policy relies on a time to collision strategy to decide when to cross and uses an additional set of priority rules:
\begin{itemize}
    \item Cars going straight or turning right on the main street have priority
    \item Pedestrians have priority and the car must yield to pedestrians approaching the crosswalk and wait for them to reach the other side of the road. 
    \item Cars follow the intelligent driver model before or after reaching the intersection.
\end{itemize}
When a car is stopped at an intersection, it measures the time to collision with other vehicles. If this time to collision is above some threshold, the car crosses. The time to collision is measured as the time it takes for the other car to reach the center of the intersection assuming the car would drive with maximum acceleration. Such a design prioritizes safety and ensures that the rule-based policy is collision free under perfect observations. In addition, we analyzed the effect of using the ensemble RNN updater versus directly using the output of the perception system. 

\textbf{Safe rule-based:} The safe rule-based policy is a combination of the model checker and the rule-based policy described above. If the action given by the rule-based policy is within the set of safe actions given by the model checker, then the action is executed. Otherwise, it executes the safest action given by the model checker.

\textbf{RL policy:} The reinforcement learning policy is trained using Deep Q Learning on the canonical scenario and is augmented using scene decomposition. The reward function assigns \num{-1} to collisions and \num{+1} for reaching the goal. The same network structure is used as for the safe-RL policy. The exact parameters used for the training are available in our code base.

Finally, we will refer to our approach combining the belief updater, model checker, RL policy, and scene decomposition as the safe-RL policy.

\begin{figure}
    \centering
    \includegraphics[width=\columnwidth]{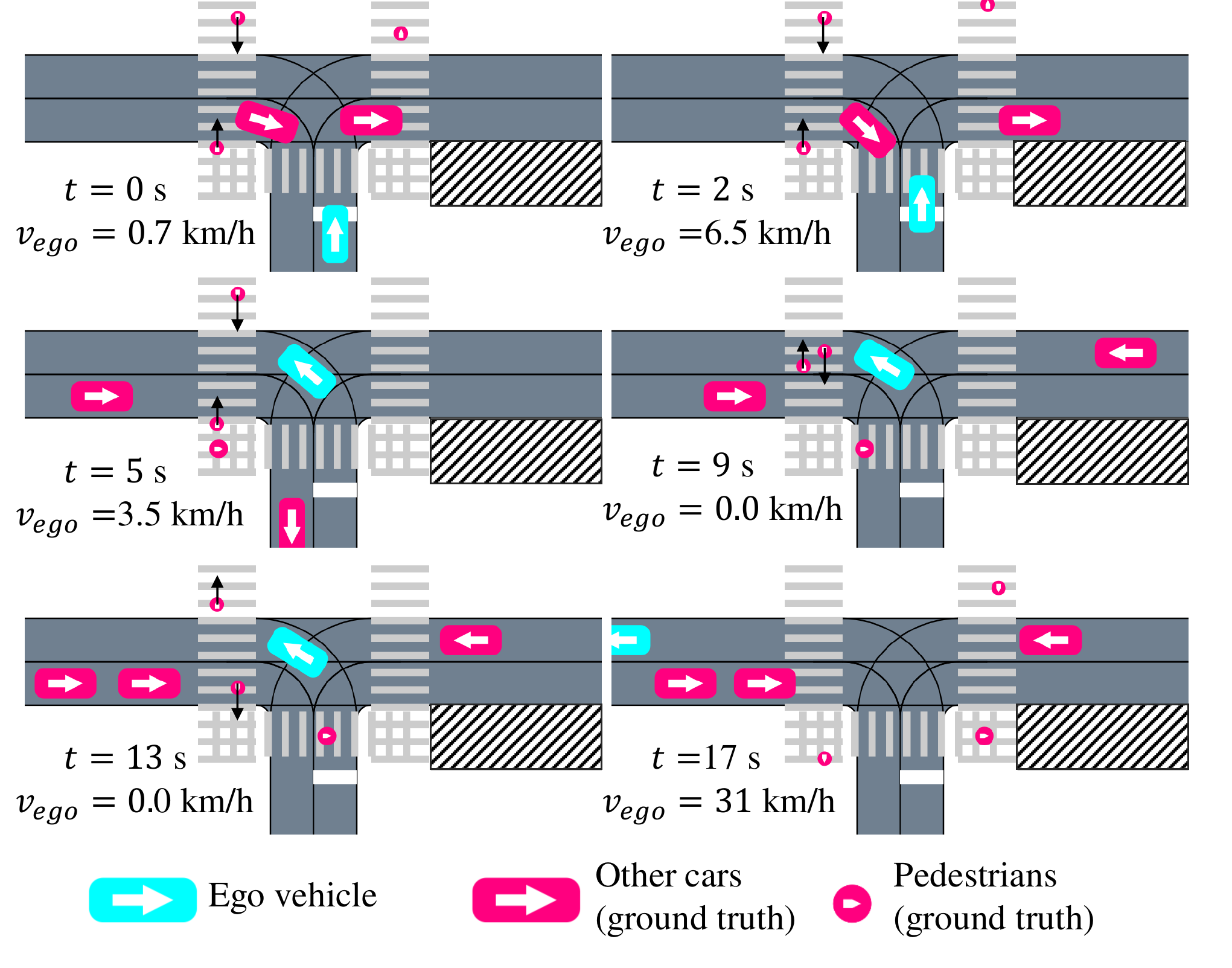}
    \caption{Example of a trajectory where the ego car executes our algorithm in scenario e). The car first engages in the intersection since no incoming cars present a risk. Then the vehicle waits for two pedestrians to cross the street. Once the pedestrians are done crossing, the autonomous vehicle pursues its course at higher speed. }
    \label{fig:trajectory}
\end{figure}

\begin{figure}[!h]
    \centering
\begin{tikzpicture}[]
\begin{groupplot}[
group style={horizontal sep=1cm, vertical sep=1.2cm, group size=1 by 2},
]
\nextgroupplot [ylabel = {Steps}, ybar=0pt, bar width=5pt, xtick=data, 
symbolic x coords={a,b,c,d,e},, ymajorgrids, ymin = {0},
legend pos = north west,
typeset ticklabels with strut,
height = 6cm,
width=8cm
]
\addplot+ [
error bars/.cd, 
x dir=both, x explicit, y dir=both, y explicit]
table [
x error plus=ex+, x error minus=ex-, y error plus=ey+, y error minus=ey-
] {
x y ex+ ex- ey+ ey-
a 97.191 0.0 0.0 0.4952374299 0.4952374299
b 163.405 0.0 0.0 0.8185292221952664 0.8185292221952664
c 92.689 0.0 0.0 0.4421718488 0.4421718488
d 122.59 0.0 0.0 0.4596525540780454 0.4596525540780454
e 136.627 0.0 0.0 0.9419082643 0.9419082643
};
\addlegendentry{RL updater}
\addplot+ [
, error bars/.cd, 
x dir=both, x explicit, y dir=both, y explicit]
table [
x error plus=ex+, x error minus=ex-, y error plus=ey+, y error minus=ey-
] {
x y ex+ ex- ey+ ey-
a 115.589 0.0 0.0 0.4710533932 0.4710533932
b 109.622 0.0 0.0 0.5507169879 0.5507169879
c 125.609 0.0 0.0 0.5260076958 0.5260076958
d 103.58 0.0 0.0 0.1941603931 0.1941603931
e 365.301 0.0 0.0 0.8106228156 0.8106228156
};
\addlegendentry{Rule-based no updater}
\addplot+ [
, error bars/.cd, 
x dir=both, x explicit, y dir=both, y explicit]
table [
x error plus=ex+, x error minus=ex-, y error plus=ey+, y error minus=ey-
] {
x y ex+ ex- ey+ ey-
a 108.632 0.0 0.0 0.33301773382 0.3301773382
b 122.906 0.0 0.0 0.7497327628 0.7497327628
c 122.312 0.0 0.0 0.4410185517 0.4410185517
d 100.379 0.0 0.0 0.1984593712 0.1984593712
e 331.513 0.0 0.0 0.9563173772 0.9563173772
};
\addlegendentry{Rule-based updater}
\addplot+ [
, error bars/.cd, 
x dir=both, x explicit, y dir=both, y explicit]
table [
x error plus=ex+, x error minus=ex-, y error plus=ey+, y error minus=ey-
] {
x y ex+ ex- ey+ ey-
a 124.945 0.0 0.0 0.2679884612 0.2679884612
b 98.019 0.0 0.0 0.2694384178 0.2694384178
c 122.376 0.0 0.0 0.49652855209999996 0.49652855209999996
d 107.575 0.0 0.0 0.21179966149999999 0.21179966149999999
e 172.778 0.0 0.0 1.083850268 1.083850268
};
\addlegendentry{Safe-RL updater}
\addplot+ [
, error bars/.cd, 
x dir=both, x explicit, y dir=both, y explicit]
table [
x error plus=ex+, x error minus=ex-, y error plus=ey+, y error minus=ey-
] {
x y ex+ ex- ey+ ey-
a 105.649 0.0 0.0 0.16985178690000002 0.16985178690000002
b 101.658 0.0 0.0 0.14514672290882483 0.14514672290882483
c 129.632 0.0 0.0 0.41289264699999995 0.41289264699999995
d 120.862 0.0 0.0 0.2575815285 0.2575815285
e 258.411 0.0 0.0 1.337956029 1.337956029
};
\addlegendentry{Safe-Rule-based updater}
\nextgroupplot [ylabel = {Collisions}, xlabel = {Scenario}, ybar=0pt, bar width=5pt, xtick=data, 
symbolic x coords={a,b,c,d,e},, ymajorgrids, ymin = {0},
legend pos = north east,
typeset ticklabels with strut,
height = 4cm,
width=8cm
]
\addplot+ [
, error bars/.cd, 
x dir=both, x explicit, y dir=both, y explicit]
table [
x error plus=ex+, x error minus=ex-, y error plus=ey+, y error minus=ey-
] {
x y ex+ ex- ey+ ey-
a 0.018 0.0 0.0 0.00133017644 0.00133017644
b 0.041 0.0 0.0 0.0019838941090279585 0.0019838941090279585
c 0.045 0.0 0.0 0.00207407854 0.00207407854
d 0.002 0.0 0.0 0.0004469897088298563 0.0004469897088298563
e 0.003 0.0 0.0 0.00054717401 0.00054717401
};
\addlegendentry{RL updater}
\addplot+ [
, error bars/.cd, 
x dir=both, x explicit, y dir=both, y explicit]
table [
x error plus=ex+, x error minus=ex-, y error plus=ey+, y error minus=ey-
] {
x y ex+ ex- ey+ ey-
a 0.008 0.0 0.0 0.0008912880500000001 0.0008912880500000001
b 0.015 0.0 0.0 0.0012161328 0.0012161328
c 0.0 0.0 0.0 0.0 0.0
d 0.0 0.0 0.0 0.0 0.0
e 0.001 0.0 0.0 0.00031622777 0.00031622777
};
\addlegendentry{Rule-based no updater}
\addplot+ [
, error bars/.cd, 
x dir=both, x explicit, y dir=both, y explicit]
table [
x error plus=ex+, x error minus=ex-, y error plus=ey+, y error minus=ey-
] {
x y ex+ ex- ey+ ey-
a 0.0 0.0 0.0 0.0 0.0
b 0.005 0.0 0.0 0.0007056897299999999 0.0007056897299999999
c 0.0 0.0 0.0 0.0 0.0
d 0.0 0.0 0.0 0.0 0.0
e 0.0 0.0 0.0 0.0 0.0
};
\addlegendentry{Rule-based updater}
\addplot+ [
, error bars/.cd, 
x dir=both, x explicit, y dir=both, y explicit]
table [
x error plus=ex+, x error minus=ex-, y error plus=ey+, y error minus=ey-
] {
x y ex+ ex- ey+ ey-
a 0.0 0.0 0.0 0.0 0.0
b 0.0 0.0 0.0 0.0 0.0
c 0.0 0.0 0.0 0.0 0.0
d 0.0 0.0 0.0 0.0 0.0
e 0.0 0.0 0.0 0.0 0.0
};
\addlegendentry{Safe-RL updater}
\addplot+ [
, error bars/.cd, 
x dir=both, x explicit, y dir=both, y explicit]
table [
x error plus=ex+, x error minus=ex-, y error plus=ey+, y error minus=ey-
] {
x y ex+ ex- ey+ ey-
a 0.0 0.0 0.0 0.0 0.0
b 0.0 0.0 0.0 0.0 0.0
c 0.0 0.0 0.0 0.0 0.0
d 0.0 0.0 0.0 0.0 0.0
e 0.0 0.0 0.0 0.0 0.0
};
\addlegendentry{Safe-Rule-based updater}
\legend{}
\end{groupplot}
\end{tikzpicture}
    \caption{Average number of steps to reach the goal (one step is \SI{0.1}{\second}), and average number of collisions of the different policies in the five evaluation scenarios. The error bar represents the standard error. The masked policies presented zero collisions across the \num{10000} simulations.}
    \label{fig:barplots}
\end{figure}

\subsection{Results and Discussion}

 In scenario a, the rule-based policy, not relying on a belief updater, has more collisions than the one using the updater. It receives a noisy observation of the true state which eventually leads to bad decisions. We can see that although the RL policy uses the belief updater, it is more aggressive and results in more collisions than the rule-based policy. In scenario b, the belief updater also reduces the number of collisions since the policy is aware of the potential presence of the car in the occluded area. Our safe-RL algorithm outperforms all the other approaches on this scenario. Our safe-RL algorithm did not cause any collisions for a comparable performance.
 
 In scenario c and d, the RL policy results in the most collisions and is the fastest to cross. Using RL in such an environment allows to take advantage of situations where the other car yields to the pedestrian. The other policies have similar performances. Since the car yielding to the pedestrian slows down, the time to collision decreases and the rule-based policy is also able to perform efficiently.
 
 In scenario e, the most complex, rule-based policies are very inefficient. In contrast, RL based policies perform very efficiently (twice as fast for RL vs rule-based). We can see that our Safe-RL approach performs comparably as the standard RL policy while being much safer. An example of a trajectory resulting from executing our algorithm in this scenario is illustrated in \cref{fig:trajectory}. The vehicle exhibits a safe and intelligent behavior. It positions itself strategically in the middle of the intersection, to leave as soon as the two pedestrian on the left are done crossing.



The simulation results highlight several important points that should be taken into consideration when designing a decision making algorithm for urban intersections. First, the belief updater plays an important role in making the algorithm robust to perception noise and aware of occluded areas. Secondly, using a pure reinforcement learning technique does not enable the agent to act more safely than when using simple rule-based methods. Moreover, we can see from the results on scenarios a, b, c, and d, which involves only one or two other traffic participants, that the efficiency of all the policies are very similar. However, when the scenario becomes more complex, with a flow of cars and pedestrians, the benefits of using reinforcement learning over a rule-based method become clear. As the set of possible situations drastically increases with the number of agents, engineering a set of rule to represent the policy is very challenging. Our safe-RL algorithm is able to find a safe policy that outperforms rule-based approaches. The presence of the model checker makes our algorithm more suitable than pure RL methods by enforcing safety constraints. 


\section{CONCLUSIONS}

This paper presented a decision making framework for autonomously navigating urban intersections. 
We introduced a learned belief updater that uses an ensemble of RNNs to estimate the location of vehicles behind obstacles and is robust to perception errors. We improved upon pure reinforcement learning methods by using a model checker to enforce safety guarantees. Finally, through a scene decomposition method, we demonstrated how to efficiently scale the algorithm to scenarios with multiple cars and pedestrians. We empirically demonstrated that our method provides safe and efficient decisions even in complex scenarios.

Future work includes applying our methodology to different situations such as highway merging and crowded driving environments. Since we demonstrated how to learn the belief updater with a synthetic dataset, a line of future work could be to use real data from the perception system of the ego vehicle. Finally, estimating the intentions of drivers has been investigated in previous works and could be useful for decision making~\cite{hubmann2018, bouton2017}.







\printbibliography

\end{document}